\documentclass[conference]{IEEEtran}
\IEEEoverridecommandlockouts
\usepackage{cite}
\usepackage{amsmath,amssymb,amsfonts}
\usepackage{algorithmic}
\usepackage{graphicx}
\usepackage{textcomp}
\usepackage{xcolor}
\usepackage{subfig}
\usepackage{algorithm}
\usepackage{hyperref}

\def\BibTeX{{\rm B\kern-.05em{\sc i\kern-.025em b}\kern-.08em
    T\kern-.1667em\lower.7ex\hbox{E}\kern-.125emX}}
\begin{document}

\title{SVD-PINNs: Transfer Learning of Physics-Informed Neural Networks via Singular 
Value Decomposition}

\author{\IEEEauthorblockN{Yihang Gao}
\IEEEauthorblockA{Department of Mathematics \\
The University of Hong Kong\\
Hong Kong SAR\\
gaoyh@connect.hku.hk}
\and
\IEEEauthorblockN{Ka Chun Cheung}
\IEEEauthorblockA{
Department of Mathematics \\
Hong Kong Baptist University and NVIDIA \\
Hong Kong SAR \\
chcheung@nvidia.com}
\and
\IEEEauthorblockN{Michael K. Ng}
\IEEEauthorblockA{Institute of Data Science and \\
Department of Mathematics \\
The University of Hong Kong\\
Hong Kong SAR \\
mng@maths.hku.hk}
}

\maketitle

\begin{abstract}
Physics-informed neural networks (PINNs) have attracted significant attention for solving partial differential equations (PDEs) in recent years because they alleviate the curse of dimensionality that appears in traditional methods. However, the most disadvantage of PINNs is that one neural network corresponds to one PDE. In practice, we usually need to solve a class of PDEs, not just one. With the explosive growth of deep learning, many useful techniques in general deep learning tasks are also suitable for PINNs. Transfer learning methods may reduce the cost for PINNs in solving a class of PDEs. In this paper, we proposed a transfer learning method of PINNs via keeping singular vectors and optimizing singular values (namely SVD-PINNs). Numerical experiments on high dimensional PDEs ($10$-d linear parabolic equations and $10$-d Allen-Cahn equations) show that SVD-PINNs work for solving a class of PDEs with different but close right-hand-side functions. 
\end{abstract}

\begin{IEEEkeywords}
Physics Informed Neural Networks, Transfer Learning, Singular Value Decomposition
\end{IEEEkeywords}

\section{Introduction}
Deep learning methods are widely studied empirically and theoretically, in computer vision \cite{krizhevsky2012imagenet}, natural language processing \cite{vinyals2015grammar} and healthcare \cite{miotto2018deep}, etc. Recently, researchers began to focus on solving complicated scientific computing problems by deep learning techniques, e.g., forward and inverse problems of PDEs \cite{raissi2019physics}, uncertainty quantification \cite{yang2019adversarial, zhang2019quantifying} as well as solving large-scale linear systems \cite{gu2022deep}.

Following some pioneering works of solving PDEs by deep neural networks \cite{psichogios1992hybrid, lagaris1998artificial}, Raissi, Perdikaris and Karniadakis \cite{raissi2019physics} proposed the physics-informed neural networks which efficiently solve high-dimensional forward and inverse PDEs problems by leveraging interior and initial/boundary conditions. Later, PINNs is extended to solve some more complicated and specific scientific computing problems, e.g., fractional PDEs \cite{pang2019fpinns}, stochastic PDEs \cite{yang2019adversarial, chen2021learning, zhang2019quantifying, zhang2020learning} and high dimensional problems \cite{weinan2021algorithms}. Yang and Perdikaris \cite{yang2019adversarial} proposed UQPINNs to do the uncertainty quantification by using GANs to measure the data distribution (uncertainty) and the PINNs term as a regularization for physical constraints. Besides introducing physical information by residuals in the vanilla PINNs, classical methods bring ideas to model deep networks. Sirignano and Spiliopoulos \cite{sirignano2018dgm} developed the deep Galerkin method with neural networks as base functions due to the universal approximation capabilities of deep neural networks.

Some theoretical and empirical investigations contribute to understanding and improving the performances of PINNs. Shin, Darbon and Karniadakis \cite{shin2020convergence} introduced Hölder regularization terms in the loss function for PINNs and analyzed the convergence of the loss in terms of the number of training data. Under Hölder's continuity, the empirical loss with a finite number of training data is close to the exact loss. Moreover, the accuracy of the solution is also guaranteed for second-order elliptic equations under the maximal principle.  Jagtap, Kawaguchi, and Karniadakis \cite{jagtap2020adaptive} introduced learnable parameters in the activation functions of neural networks to adaptively control learning rates and weights for each node. In \cite{wang2022and}, Wang, Yu, and Perdikaris analyzed the training dynamics of PINNs using neural tangent kernel theory and adaptively assigned weights of the loss to accelerate the convergence of PINNs.

However, there are still a bunch of unsolved works for PINNs to be investigated. For example, how to automatically determine hyperparameters that balance interior and initial/boundary conditions? To the best of our knowledge, we tune the hyperparameters mostly by trials and experience. Moreover, we can only solve a single PDE by one neural network and we are required to retrain and store the model for another one even if they have some similarities. For example, if several PDEs have the same differential operators but different right-hand side functions, can we reduce the computation and storage for solving these PDEs based on their similarities and shared information? In this paper, we aim to solve the latter problem for PINNs by transfer learning methods. 

\subsection{Transfer Learning}
Transfer learning is a machine learning problem that aims to gain some information from one problem and apply it to different but related problems for lower costs \cite{pan2009survey}. For deep neural networks, transfer learning may generalize better and avoid overfitting \cite{yosinski2014transferable}. Parameters in front layers are usually used for extracting some fundamental information and thus are applicable for various related tasks. Therefore, a straightforward strategy is fixing the front layers and only optimizing the top layers in the current task. Moreover, previously trained parameters sometimes are good initialization for current models.

A natural question is whether transfer learning is valid for training PINNs. Specifically, for a class of PDEs whose differential operators are the same but their right-hand side functions are different, transfer learning may help to train PINNs at lower costs. The idea of transfer learning for training PINNs was studied in \cite{desai2021one}. Parameters of hidden layers are frozen and only the parameters for the output layer are trainable. Experimental results in \cite{desai2021one} show that transfer learning works well and reduces the storage of PINNs. A transfer neuroevolutionary algorithm was proposed recently in \cite{wong2021can}, which is computationally more efficient than the original neuroevolutionary algorithm in training PINNs. Unlike some popular transfer learning methods, it does not strictly freeze any of the model parameters, allowing the learning algorithm to adaptively transfer online.

\subsection{The Contribution}
In this paper, we proposed a singular-values-based transfer learning method of PINNs for solving a class of PDEs, namely SVD-PINNs. Compared with works in \cite{desai2021one}, the proposed SVD-PINNs freeze the bases for the parameters matrix of the hidden layer and singular values are trainable. Numerical experiments on high-dimensional PDEs ($10$-d linear parabolic equations and $10$-d Allen-Cahn equations) are conducted. We observed that the main challenge and difficulty of the proposed method is the optimization of singular values. Successful optimization of singular values contributes to better performances of SVD-PINNs than the model in \cite{desai2021one}. Conversely, 
SVD-PINNs with biased and inaccurate singular values have worse results.

The outline of the paper is given as follows. In section \ref{sec_methods}, we first briefly introduce and review PINNs and a transfer learning model. Then the SVD-PINNs method is presented for solving a class of PDEs. In section \ref{sec_experimental_results}, some numerical observations are displayed. We discuss some potential future works in section \ref{sec_conclusion_discussion}. 

\subsection{Some Potential Limitations}
However, there are some potential limitations of the SVD-PINNs that are not well studied in the paper and we left them as future works. 

Firstly, the theoretical analysis for SVD-PINNs. In section \ref{sec_methods}, we provide some intuitive explanations for designing SVD-PINNs. However, it lacks strict mathematical guarantees (e.g., convergence and generalization analysis). 

Secondly, the optimization for singular values. Gradient-based (global) methods and projections are adopted for optimizing SVD-PINNs since singular values are non-negative. However, these optimizers usually are not valid and convergent for solving constrained optimization problems, without the convexity condition. Numerical results show that successful optimization of singular values for SVD-PINNs performs well, while SVD-PINNs with inaccurate singular values have large relative errors to solutions.

\subsection{Notations}
Throughout the paper, we denote $\|\cdot\|_2$ as the Euclidean norm for vectors and the corresponding $2$-norm for matrices. We use boldface capital and lowercase letters to denote matrices and vectors respectively. Singular value decomposition (SVD) always exists for any real matrix $\mathbf{A} \in \mathbb{R}^{m}$, i.e.,
there exist real unitary matrices $\mathbf{U}$ and $\mathbf{V}$, and non-negative diagonal matrix $\mathbf{D}$, such that
\begin{equation*}
    \mathbf{A} = \mathbf{U} * \mathbf{D} * \mathbf{V}^{\mathrm{T}},
\end{equation*}
where diagonals of $\mathbf{D} = \text{diag}(\sigma_1, \cdots, \sigma_m)$ are singular values of $\mathbf{A}$. The relative error is defined as
\begin{equation*}
    \text{err}(\{\Tilde{y}_{i}\}_{i};\{\mathbf{y}_{i}\}) = \sqrt{\frac{\sum_{i=1}^{n} (\Tilde{y}_{i} - y_i)^2}{\sum_{i=1}^{n} y_i^2}},
\end{equation*}
where $\{\Tilde{y}_{i}\}_{i}$ are the predictions and $\{\mathbf{y}_{i}\}$ are accurate values.

\section{Methods}
\label{sec_methods}

\subsection{Physics-Informed Neural Networks (PINNs)}
We first briefly review the significant work in \cite{raissi2019physics}, which opened the door for mathematical machine learning in scientific computing. 
For a given PDE
\begin{equation}
\label{pde}
    \begin{split}
        \mathcal{D}[\mathbf{u}, \mathbf{x}] & = \mathbf{f}(\mathbf{x}), \hspace{1em} \mathbf{x} \in \Gamma \subset \mathbb{R}^{d},\\
        \mathcal{B}[\mathbf{u}, \mathbf{x}] & = \mathbf{g}(\mathbf{x}), \hspace{1em} \mathbf{x} \in \partial \Gamma,
    \end{split}
\end{equation}
where $\mathcal{D}$ and $\mathcal{B}$ are differential operators in the interior and on the boundary respectively, $\Gamma$ is an open set of our interest and $\partial \Gamma$ is its boundary, we adopt a neural network $\phi(\mathbf{x};\theta)$ as a surrogate to the solution $\mathbf{u}(\mathbf{x})$. Here, $\phi(\mathbf{x};\theta)$ is a fully connected neural network parameterized by $\theta$. In this paper, we consider two-hidden-layer neural networks defined as
\begin{equation}
    \phi(\mathbf{x};\theta) = \mathbf{W}_{2} \cdot \left(\sigma \left (\mathbf{W}_{1} \cdot \sigma \left(\mathbf{W}_{0}\mathbf{x} + \mathbf{b}_{0} \right)+\mathbf{b}_{1} \right) \right) + \mathbf{b}_{2},
\end{equation}
and
\begin{equation*}
    \theta = \left\{\mathbf{W}_{2}, \mathbf{W}_{1}, \mathbf{W}_{0}, \mathbf{b}_{2}, \mathbf{b}_{1}, \mathbf{b}_{0} \right\},
\end{equation*}
with activation function $\sigma(x) = \text{ReLU}^{k}(x)$ \cite{xu2020finite} or $\sigma(x) = \text{tanh}(x)$ \cite{raissi2019physics}. With interior training samples $\{\mathbf{x}_{i}\}_{i=1}^{n_1}$ and initial/boundary samples $\{\Tilde{\mathbf{x}}_{j}\}_{j=1}^{n_2}$, the loss function of PINNs is formulated as
\begin{equation}
\label{loss_pinns}
\begin{split}
    \min_{\theta} \hspace{1em} & \nu \cdot \frac{1}{n_1}  \sum_{i=1}^{n_1} \frac{1}{2} \left \|\mathcal{D}[\phi(\mathbf{x}_{i}; \theta), \mathbf{x}_{i}] - \mathbf{f}(\mathbf{x}_{i})\right\|_2^2\\
    & + \frac{1}{n_2} \sum_{j=1}^{n_2} \frac{1}{2} \left \|\mathcal{B}[\phi(\Tilde{\mathbf{x}}_{j}; \theta), \Tilde{\mathbf{x}}_{j}] - \mathbf{g}(\Tilde{\mathbf{x}}_{j})\right\|_2^2,
\end{split}
\end{equation}
where the hyperparameter $\nu>0$ balances the interior and initial/boundary conditions.

The method above solves a single PDE, which implies that one neural network corresponds to one PDE, even if several PDEs have similarities and shared information. DeepONet \cite{lu2021learning} is capable of solving a class of PDEs by learning the differential operator from training samples. However, it requires sufficient solution data, which is usually inaccessible. 

\subsection{Transfer Learning of PINNs}
Desai, Mattheakis, Joy, Protopapas, and Roberts \cite{desai2021one} proposed a transfer learning method of PINNs for solving a class of PDEs. For a class of PDEs with the same differential operators $\mathcal{D}$ and $\mathcal{B}$ but different right-hand side functions $\{\mathbf{f}_{\epsilon}(\mathbf{x})\}_{\epsilon}$ and $\{\mathbf{g}_{\epsilon}(\mathbf{x})\}_{\epsilon}$, the corresponding approximate solution $\phi(\mathbf{x};\theta_{\epsilon})$ shares some parameters $\left\{\mathbf{W}_{1}, \mathbf{W}_{0}, \mathbf{b}_{2}, \mathbf{b}_{1}, \mathbf{b}_{0} \right\}$ but only $\mathbf{W}_2$ is trainable. More specifically, we first pretrain the model $\phi(\mathbf{x};\theta_{\epsilon})$ for a given $\epsilon$, and then parameters $\left\{\mathbf{W}_{1}, \mathbf{W}_{0}, \mathbf{b}_{2}, \mathbf{b}_{1}, \mathbf{b}_{0} \right\}$ are frozen. For other $\epsilon$, we only optimize the parameters $\mathbf{W}_2$ in the output layer to minimize (\ref{loss_pinns}). Note that the loss function for $\mathbf{W}_2$ is convex, when the differential operators $\mathcal{D}$ and $\mathcal{B}$ are linear. Therefore, efficient solvers can accurately find the optimal $\mathbf{W}_2$. However, it raises a concern for the capacity of the model, since only $\mathbf{W}_2$ is trainable.

\subsection{SVD-PINNs}
In this part, we present a novel singular-value-decomposition (SVD) 
based transfer learning of PINNs, namely SVD-PINNs. 

\subsubsection{Motivation}
It is redundant and costly to train PINNs one-by-one for a class of right-hand side functions $\{\mathbf{f}_{\epsilon}(\mathbf{x})\}_{\epsilon}$ and $\{\mathbf{g}_{\epsilon}(\mathbf{x})\}_{\epsilon}$. If the image of the solution is in $\mathbb{R}^{r}$ and both the first and the second hidden layers are of the same length $m$, then we need to store $\left(m^2 + (r + d + 1) \cdot m + r\right)$ parameters for each PINN. Therefore, techniques that make use of shared information and reduce costs are essential for solving a class of similar PDEs. When $m$ is large, the dominant storage comers from the matrix $\mathbf{W}_1 \in \mathbb{R}^{m \times m}$. 

The parameters matrix $\mathbf{W}_1$ is mostly related to the differential operators $\mathcal{D}$ and $\mathcal{B}$. Considering a linear PDE, the finite difference method is equivalent to solving a linear system
\begin{equation}
    \mathbf{A}\mathbf{v} = \mathbf{b}.
\end{equation}
Here, the matrix $\mathbf{A}$ is a discretization of the differential operator $\mathcal{D}$ with $\mathcal{B}$. The vector $\mathbf{v}$ denotes the approximate solution at discretization (grid) points, i.e., $\mathbf{v} \approx [\mathbf{u}(\mathbf{x}_i)]$ for $\mathbf{x}_i \in \mathcal{S}$, where $\mathcal{S}$ is the set of grid points. 
If we approximate the solution by neural networks $\phi(\mathbf{x};\theta)$, i.e.,
\begin{equation*}
    \mathbf{A} \phi(\mathbf{X};\theta) = \mathbf{b},
\end{equation*}
then 
\begin{equation*}
    \mathbf{A} \left(\sigma \left (\sigma \left(\mathbf{X}\mathbf{W}_{0}^{\top} \right)  \right) \cdot \mathbf{W}_{1}^{\top} \right) \cdot \mathbf{W}_{2}^{\top} = \mathbf{b},
\end{equation*}
where we omit the bias terms in the neural networks and $\mathbf{X}=[\mathbf{x}_i]$ is the collection matrix of grid points. For different right-hand functions $\mathbf{f}$ and $\mathbf{g}$, we may merely need to change $\mathbf{b}$. Therefore, the transfer learning of PINNs with training on $\mathbf{W}_2$ performs well in this sense. However, empirically, we will never achieve the optimal (exact) weights of $\mathbf{W}_{0}$ and $\mathbf{W}_{1}$. However, the estimated weights should be close enough to the desired ones. For further corrections on estimated $\mathbf{W}_{0}$ and $\mathbf{W}_{1}$ but keeping lower computational and storage costs, we freeze the base vectors of $\mathbf{W}_{1}$. Specifically, for PDEs with different right-hand side functions, the corresponding approximate solution $\phi(\mathbf{x};\theta_{\epsilon})$ may share the bases of $\mathbf{W}_1$, i.e., $\mathbf{U}$ and $\mathbf{V}$ are frozen with $\mathbf{W}_1 = \mathbf{U} * \mathbf{D} * \mathbf{V}^{\mathrm{T}}$ for different $\epsilon$. Therefore, instead of updating the whole matrix $\mathbf{W}_1$, only $m$ singular values of $\mathbf{W}_1$ are trainable. 

\subsubsection{Algorithm} 
Bases $\{\mathbf{U}, \mathbf{V}\}$ for $\mathbf{W}_1$ are frozen after pretraining and parameters $\left\{\mathbf{D}, \mathbf{W}_{2}, \mathbf{W}_{0}, \mathbf{b}_{2}, \mathbf{b}_{1}, \mathbf{b}_{0} \right\}$ are trainable, where $\mathbf{D}$ is the singular values of $\mathbf{W}_1$. The detailed algorithm for the 
singular-value-decomposition-based transfer learning of PINNs (SVD-PINNs) is shown in Algorithm \ref{alg_svd_pinns}. We adopt some state-of-the-art gradient-based first-order methods to iteratively update parameters (in lines 6 and 7 in Algorithm \ref{alg_svd_pinns}), e.g., simple gradient descent, RMSProp \cite{hinton2012rmsprop} and Adam \cite{kingma2015adam}. Numerical explorations in section \ref{sec_experimental_results} show better performances of gradient descent and RMSProp than Adam in several examples. However, we should be careful that singular values $\mathbf{D}$ should be non-negative. Therefore, the target is a constrained optimization problem. The above gradient-based (global) methods with projection are not guaranteed to converge and find the optimal solution, without the convexity condition. Gradient-based (global) methods with projection are widely used for solving constrained (nonconvex) problems in deep learning applications and achieving numerical efficiency. For example, in training Wasserstein GANs, we first transform the Lipschitzness constraints to the boundedness of parameters, and then gradient-based (global) methods with clipping are adopted \cite{arjovsky2017wasserstein}. 

\subsubsection{Advantages}
Imaging that if we aim to solve $n$ PDEs with the same differential operators but different right-hand side functions $\mathbf{f}_{\epsilon}$ and $\mathbf{g}_{\epsilon}$, the general PINNs method without transfer learning are required to optimize and store $n$ different neural networks, where the total storage is of $n \cdot \left(m^2 + (r + d + 1) \cdot m + r\right)$ parameters. For the proposed SVD-PINNs, $n \cdot \left((r + d + 2) \cdot  m + r\right) + 2m^2$ parameters are stored. The SVD-PINNs method saves much storage compared with the general PINNs method when $n \gg 1$.

Furthermore, numerical comparisons in section \ref{sec_experimental_results} show the reasonableness and effectiveness of singular-values optimization in the transfer learning of PINNs (SVD-PINNs). 

 \begin{algorithm}[t!]
 \caption{SVD-PINNs}
 \label{alg_svd_pinns}
 \begin{algorithmic}[1]
 \renewcommand{\algorithmicrequire}{\textbf{Input: samples $\{\mathbf{x}_{i}\}_{i=1}^{n_1}$ and $\{\Tilde{\mathbf{x}}_{j}\}_{j=1}^{n_2}$}}
 \renewcommand{\algorithmicensure}{\textbf{Output: $\{\theta_{\epsilon}\}_{\epsilon}$}}
 \REQUIRE in
 \ENSURE  out
 \\ \textit{Pretraining} :
  \STATE Using gradient-based methods to optimize the loss (\ref{loss_pinns}) with $\mathbf{f}_{0}$ and $\mathbf{g}_{0}$, we have the output parameters $\theta_{0}=\left\{\mathbf{W}_{2}, \mathbf{W}_{1}, \mathbf{W}_{0}, \mathbf{b}_{2}, \mathbf{b}_{1}, \mathbf{b}_{0} \right\}$. 
  \STATE Applying the SVD to $\mathbf{W}_1=\mathbf{U} * \mathbf{D} * \mathbf{V}^{\mathrm{T}}$ and $\{\mathbf{U}, \mathbf{V}\}$ are frozen for all $\theta_{\epsilon}$.  $\{\mathbf{D}, \mathbf{W}_2, \mathbf{W}_{0}, \mathbf{b}_{2}, \mathbf{b}_{1}, \mathbf{b}_{0} \}$ are trainable parameters. 
  \textit{Transfer Learning of PINNs for a given $\epsilon$ (Finetuning)}:
  \STATE $\theta_{\epsilon}(0) = \theta_{0}$.
  \FOR {$t = 1$ to $T$}
  \STATE Updating parameters $\{\mathbf{W}_2, \mathbf{W}_{0}, \mathbf{b}_{2}, \mathbf{b}_{1}, \mathbf{b}_{0} \}$.
  \STATE Updating singular values $\mathbf{D}$.
  \STATE We have the current parameters $\theta_{\epsilon}(t)$.
  \ENDFOR
 \STATE $\theta_{\epsilon} = \theta_{\epsilon}(T)$
 \RETURN $\{\theta_{\epsilon}\}_{\epsilon}$
 \end{algorithmic}
 \end{algorithm}



\begin{figure*}[t!]
  \centering
  \subfloat[Different optimizers.] {
     \label{linear_parabolic_10d_error_0.50_optimizers}     
    \includegraphics[width=0.63\textwidth]{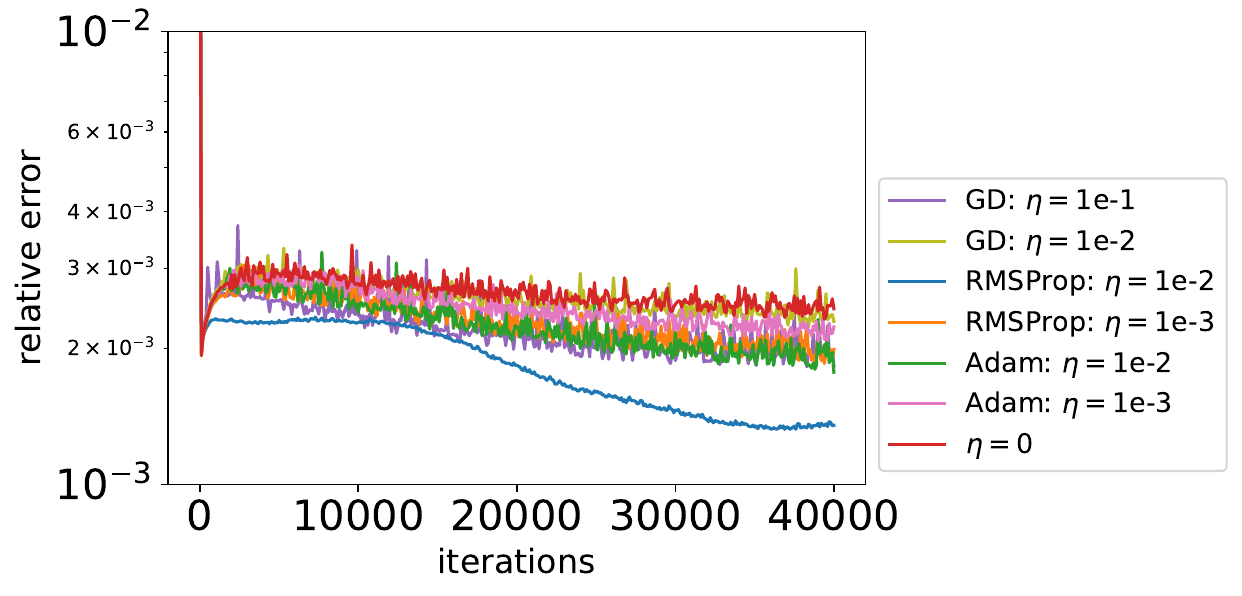}
    }\
    \subfloat[Comparisons.]{
    \label{linear_parabolic_10d_error_0.50_comp}   
    \includegraphics[width=0.63\textwidth]{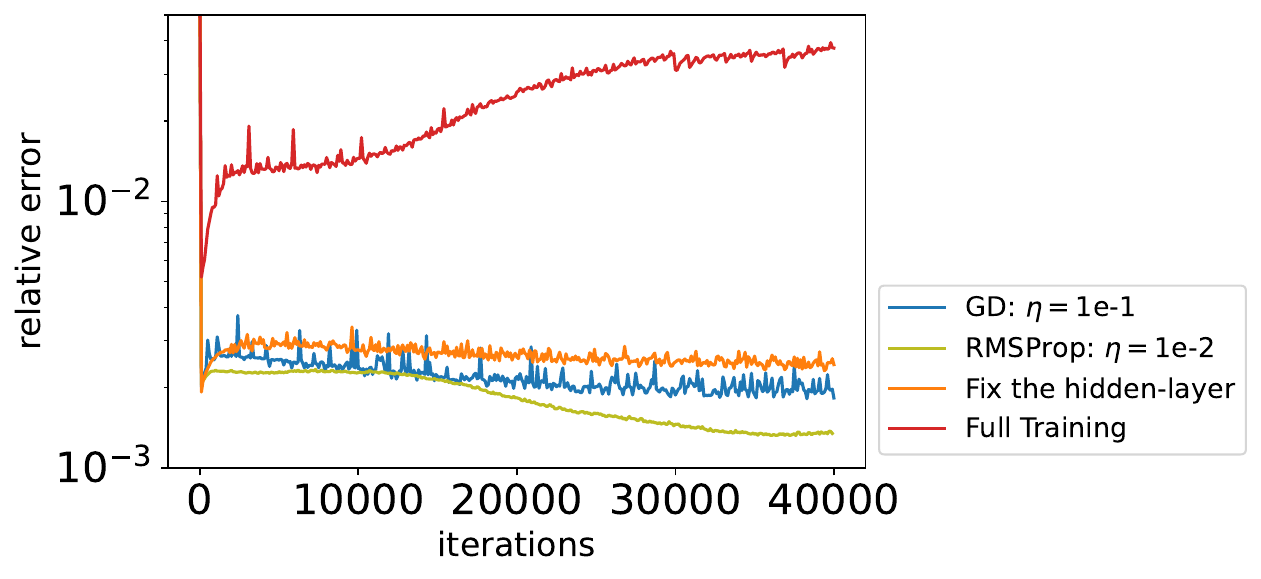}
    }\
    \caption{Trajectories of the relative error for the SVD-PINNs with different optimizers and learning rates in solving the $10$-dimensional linear parabolic  equation ($\epsilon = 0.5$).}
    \label{linear_parabolic_10d_error_0.50}
\end{figure*}

\section{Experimental Results}
\label{sec_experimental_results}
In this section, we report experimental results on transfer learning of PINNs in solving high dimensional PDEs ($10$-d Allen-Cahn equations and $10$-d hyperbolic equations) via optimizing singular values. Compared with the simple method that fixes the hidden layer, the proposed method achieves lower relative error for the solution.

All parameters are optimized by the Adam optimizer with the learning rate $1$e-3, except updating singular values in the proposed SVD-PINNs method. Two-hidden-layer and fully connected neural networks are adopted.

\subsection{Linear Parabolic Equations}
We test the proposed SVD-PINNs in solving the following $10$-dimensional linear parabolic equations:
\begin{equation}
\begin{split}
    & \frac{\partial u}{\partial t} (t,\mathbf{x}) - \nabla_{\mathbf{x}} \cdot \left(a(\mathbf{x}) \nabla_{\mathbf{x}}u(t,\mathbf{x}) \right) = f_{\epsilon}(t,\mathbf{x}), \text{ in } (0,1) \cup \Omega,\\
    & u(t,\mathbf{x}) = g_{\epsilon}(t,\mathbf{x}), \text{ on } (0,1) \cup \partial \Omega,\\
    & u(0, \mathbf{x}) = h_{\epsilon}(\mathbf{x}), \text{ in } \Omega,
\end{split}
\end{equation}
where $a(\mathbf{x}) = 1 + \frac{1}{2}\|\mathbf{x}\|_2$, $\Omega=\{\mathbf{x}: \|\mathbf{x}\|_2 < 1\}$ is the region of our interest for the spatial domain and $\partial \Omega=\{\mathbf{x}: \|\mathbf{x}\|_2 = 1\}$ is its boundary. Here, right-hand sides of the PDE (i.e., $f_{\epsilon}$, $g_{\epsilon}$ and $h_{\epsilon}$) are set by the exact solution
\begin{equation}
    u_{\epsilon}(t,\mathbf{x}) = \exp \left(\|\mathbf{x}\|_2 \cdot \sqrt{1-t} + \epsilon \cdot (1-t) \right).
\end{equation}
Note that $u_{\epsilon}$, $f_{\epsilon}$, $g_{\epsilon}$ and $h_{\epsilon}$ are differential (and thus continuous) with respect to $\epsilon$. We first pretrain the model on the PDE with $\epsilon=0$, and then some transfer learning strategies are adopted in solving PDEs with different $\epsilon$ (i.e., $\epsilon=0.5$ and $\epsilon=2$).

\begin{figure*}[t!]
  \centering
  \subfloat[Different optimizers] {
     \label{linear_parabolic_10d_error_2.00_optimizers}     
    \includegraphics[width=0.63\textwidth]{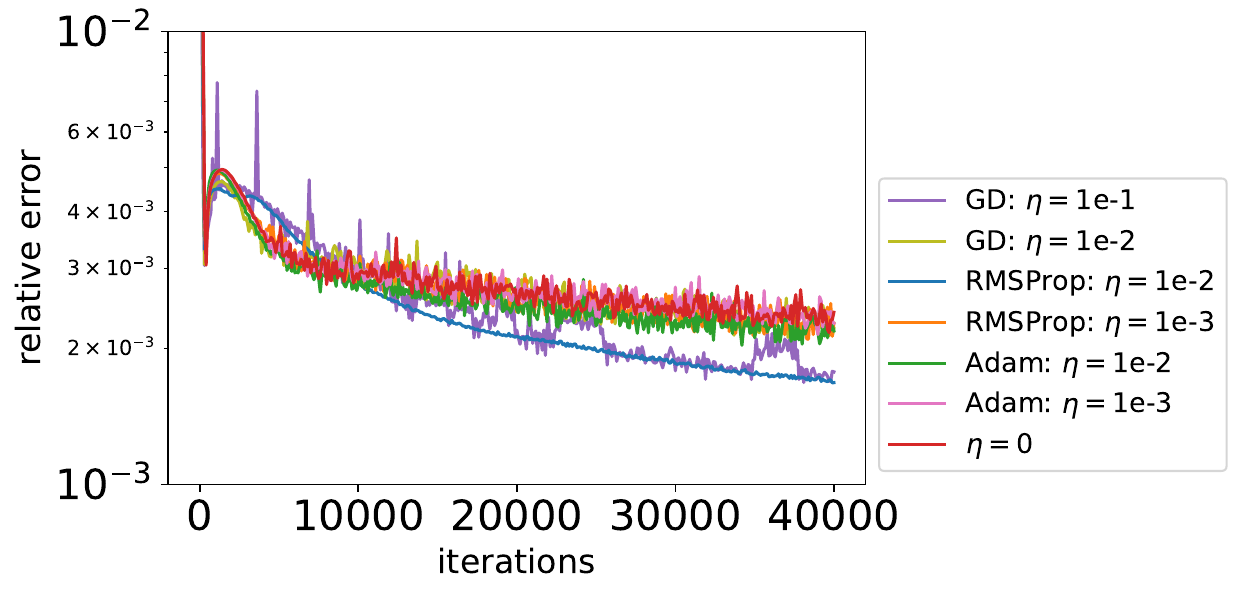}
    }\
    \subfloat[Comparisons.]{
    \label{linear_parabolic_10d_error_2.00_comp}   
    \includegraphics[width=0.63\textwidth]{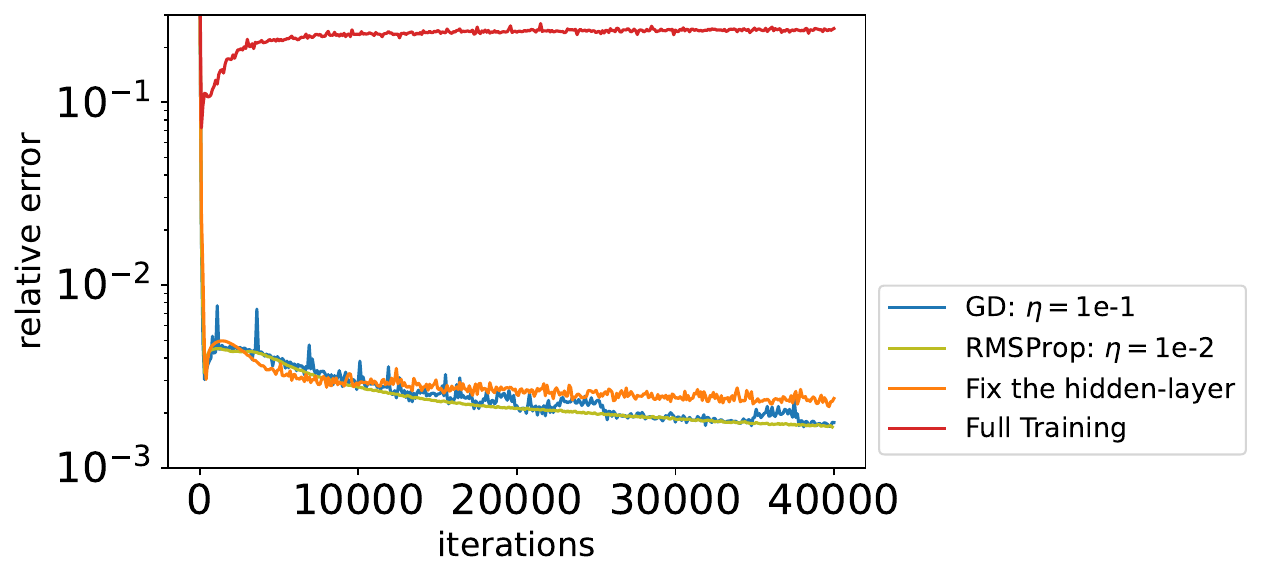}
    }\
    \caption{Trajectories of the relative error for the SVD-PINNs with different optimizers and learning rates in solving the $10$-dimensional linear parabolic  equation ($\epsilon = 2$).}
    \label{linear_parabolic_10d_error_2.00}
\end{figure*}

Figures \ref{linear_parabolic_10d_error_0.50} and \ref{linear_parabolic_10d_error_2.00} show results of SVD-PINNs in solving linear parabolic equations with $\epsilon=0.5$ and $\epsilon=2$ respectively. We try different state-of-the-art optimizers in deep learning, e.g., simple gradient descent (GD), RMSProp, and Adam, in optimizing singular values of the parameters matrix $\mathbf{W}_1$. We observe from curves in Figures \ref{linear_parabolic_10d_error_0.50_optimizers} and \ref{linear_parabolic_10d_error_2.00_optimizers} that GD with learning rate $\eta=1$e-1 and RMSProp with $\eta = 1$e-2 achieve the lowest relative error of the solution. Moreover, SVD-PINNs with other optimizers perform comparatively or a little bit better than the simple transfer learning method that fixes the parameters $\mathbf{W}_1$ (i.e., $\eta=0$). It further implies that we should focus on the optimization of singular values, since inappropriate optimization may lead to worse results. Note that we apply the gradient-based methods for singular values and then directly do clipping (projection) to guarantee the non-negative definiteness. However, as mentioned before, singular values are non-negative, and thus optimizing SVD-PINNs is actually a constrained optimization problem. Strictly speaking, gradient-based methods with clipping may not be suitable for SVD-PINNs. Therefore, a more practical and reasonable method for optimizing SVD-PINNs may further enhance the performance of SVD-PINNs and reduce the relative error. We also compare the proposed method and the simple transfer learning method with full training that updates all parameters. As are shown in Figures \ref{linear_parabolic_10d_error_0.50_comp} and \ref{linear_parabolic_10d_error_2.00_comp}, full training (i.e., training all parameters simultaneously) is sensitive to learning rates and achieves larger relative error while transfer learning methods are relatively more stable. It is consistent with the fact that transfer learning may stabilize the training of parameters and generalize better.

\begin{figure*}[t!]
  \centering
  \subfloat[Different optimizers] {
     \label{allen_cahn_10d_error_0.50_optimizers}     
    \includegraphics[width=0.63\textwidth]{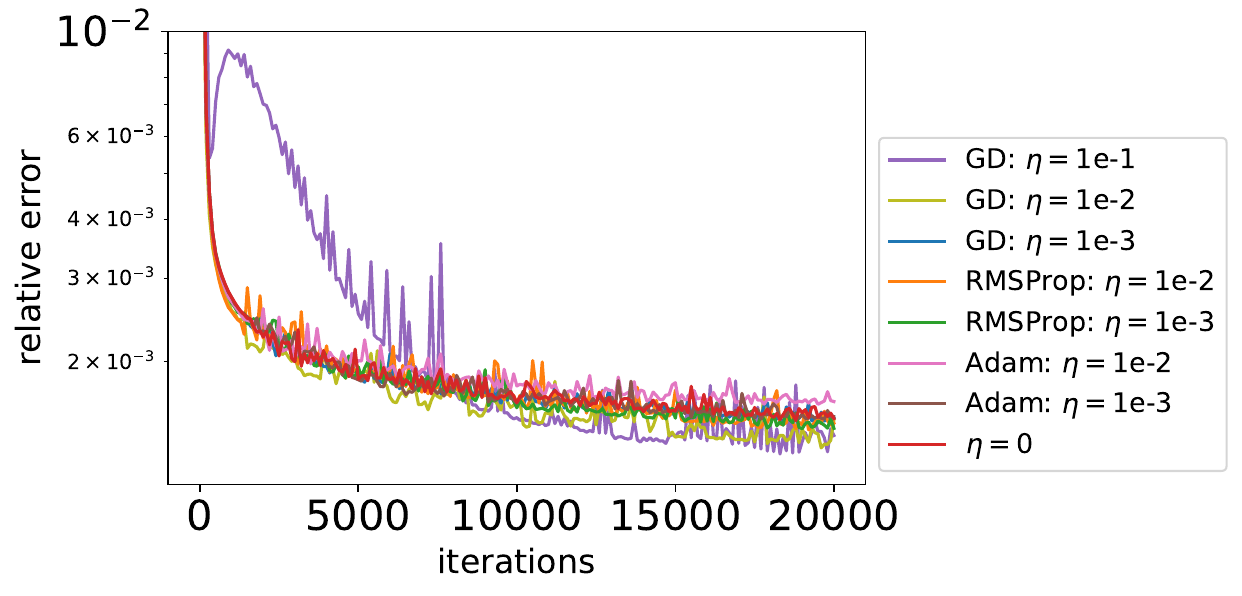}
    }\
    \subfloat[Comparisons.]{
    \label{allen_cahn_10d_error_0.50_comp}   
    \includegraphics[width=0.63\textwidth]{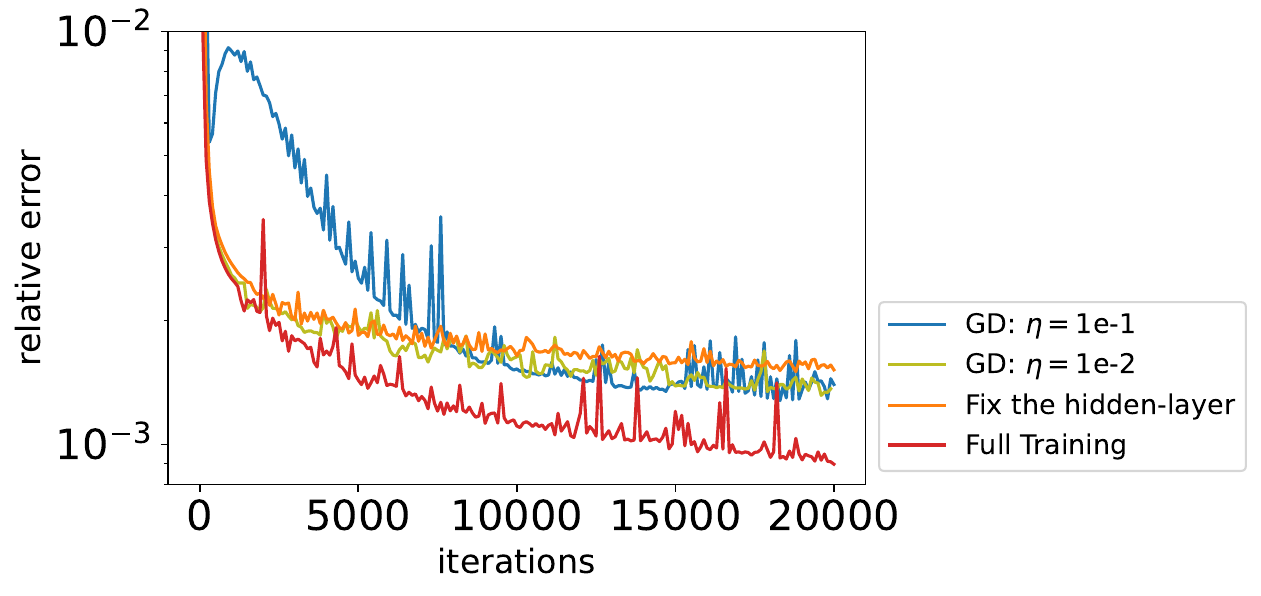}
    }\
    \caption{Trajectories of the relative error for the SVD-PINNs with different optimizers and learning rates in solving the $10$-dimensional Allen-Cahn equation ($\epsilon = 0.5$).}
    \label{allen_cahn_10d_error_0.50}
\end{figure*}

\subsection{Allen-Cahn Equations}
In this section, we consider the following $10$-dimensional nonlinear parabolic (Allen-Cahn) equation:

\begin{equation}
\begin{split}
    & \frac{\partial u}{\partial t} (t,\mathbf{x}) - \Delta_{\mathbf{x}} u(t,\mathbf{x}) - u(t,\mathbf{x}) + u^3(t,\mathbf{x})= f_{\epsilon}(t,\mathbf{x}),\\
    & \hspace{18em} \text{ in } (0,1) \cup \Omega,\\
    &u(t,\mathbf{x}) = g_{\epsilon}(t,\mathbf{x}), \text{ on } (0,1) \cup \partial \Omega\\
    & u(0, \mathbf{x}) = h_{\epsilon}(\mathbf{x}), \text{ in } \Omega, \\
\end{split}
\end{equation}
with $\Omega=\{\mathbf{x}: \|\mathbf{x}\|_2 < 1\}$. Here, right-hand sides of the PDE (i.e., $f_{\epsilon}$, $g_{\epsilon}$ and $h_{\epsilon}$) are set by the exact solution
\begin{equation}
\begin{split}
    u_{\epsilon}(t,\mathbf{x}) = \exp \left(-t \right) \cdot \left( \sin\left(\frac{\pi}{2}\left(1-\|\mathbf{x}\|_2\right)^{2.5}\right) \right. \\
    \left.+ \epsilon \cdot \sin \left( \frac{\pi}{2}\left(1-\|\mathbf{x}\|_2\right)\right)\right).
\end{split}
\end{equation}
Similarly, $u_{\epsilon}$, $f_{\epsilon}$, $g_{\epsilon}$ and $h_{\epsilon}$ are differentiable (and thus continuous) with respect to $\epsilon$. We first pretrain the model on the Allen-Cahn equation with $\epsilon=0$, and then apply some transfer learning strategies to solve PDEs with $\epsilon=0.5$, $\epsilon=2$ and $\epsilon=50$.

\begin{figure*}[t!]
  \centering
  \subfloat[Different optimizers] {
     \label{allen_cahn_10d_error_2.00_optimizers}     
    \includegraphics[width=0.63\textwidth]{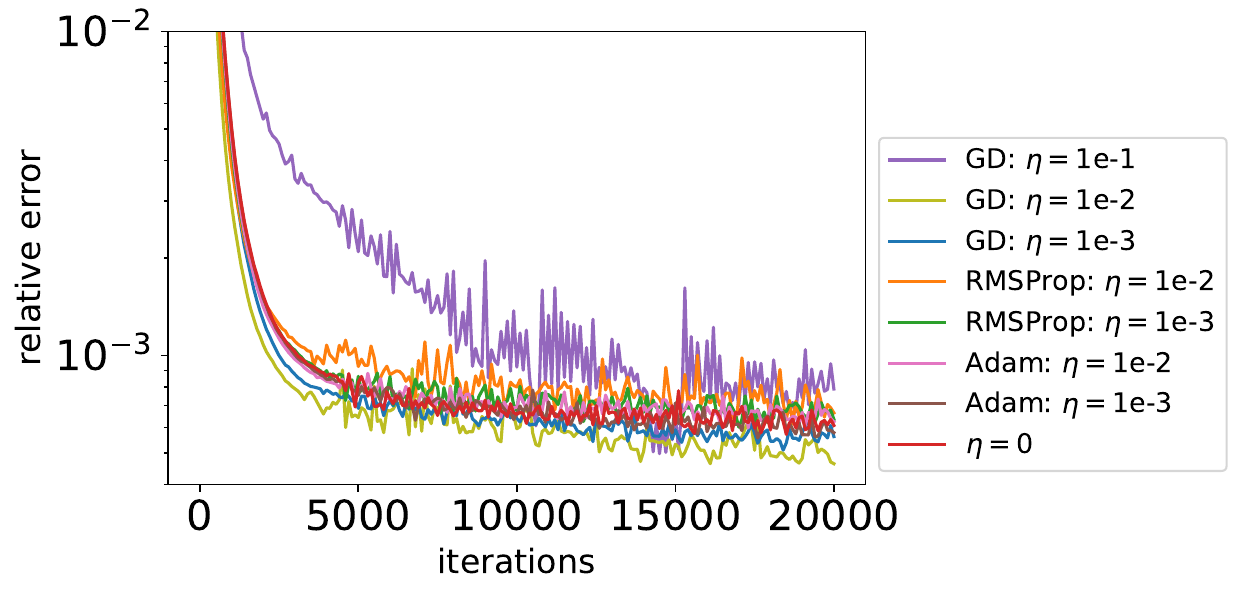}
    }\
    \subfloat[Comparisons.]{
    \label{allen_cahn_10d_error_2.00_comp}   
    \includegraphics[width=0.63\textwidth]{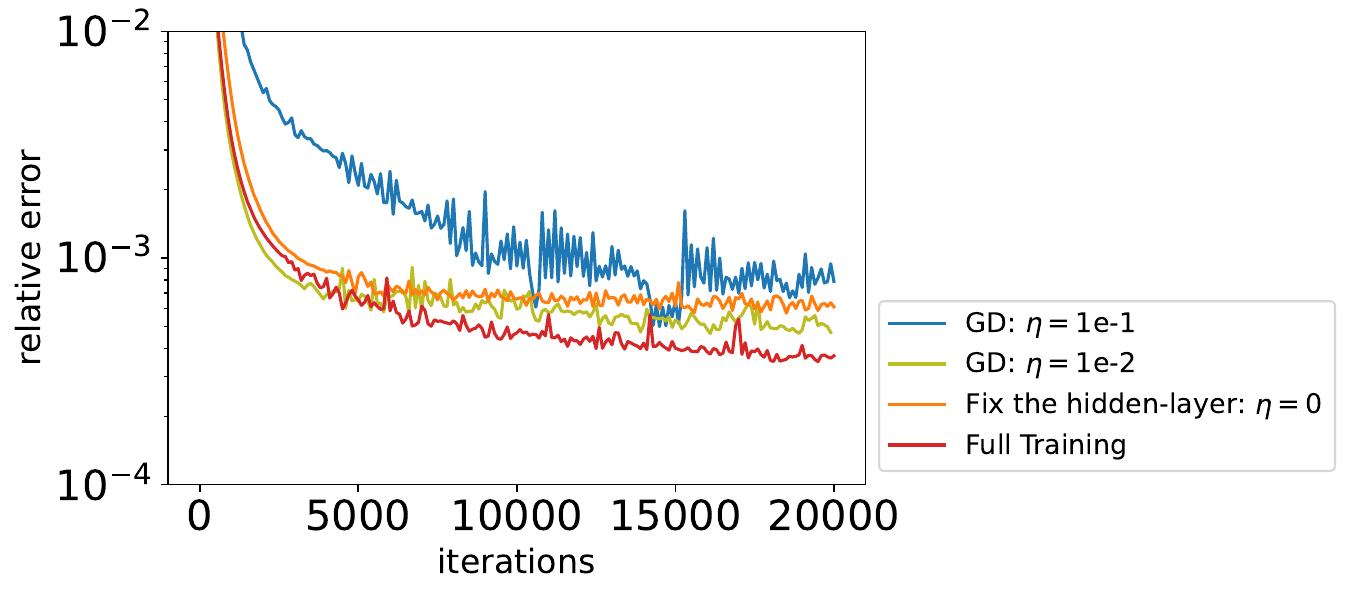}
    }\
    \caption{Trajectories of the relative error for the SVD-PINNs with different optimizers and learning rates in solving the $10$-dimensional Allen-Cahn equation ($\epsilon = 2$).}
    \label{allen_cahn_10d_error_2.00}
\end{figure*}

\begin{figure*}[t!]
    \centering
    \includegraphics[width=0.63 \textwidth]{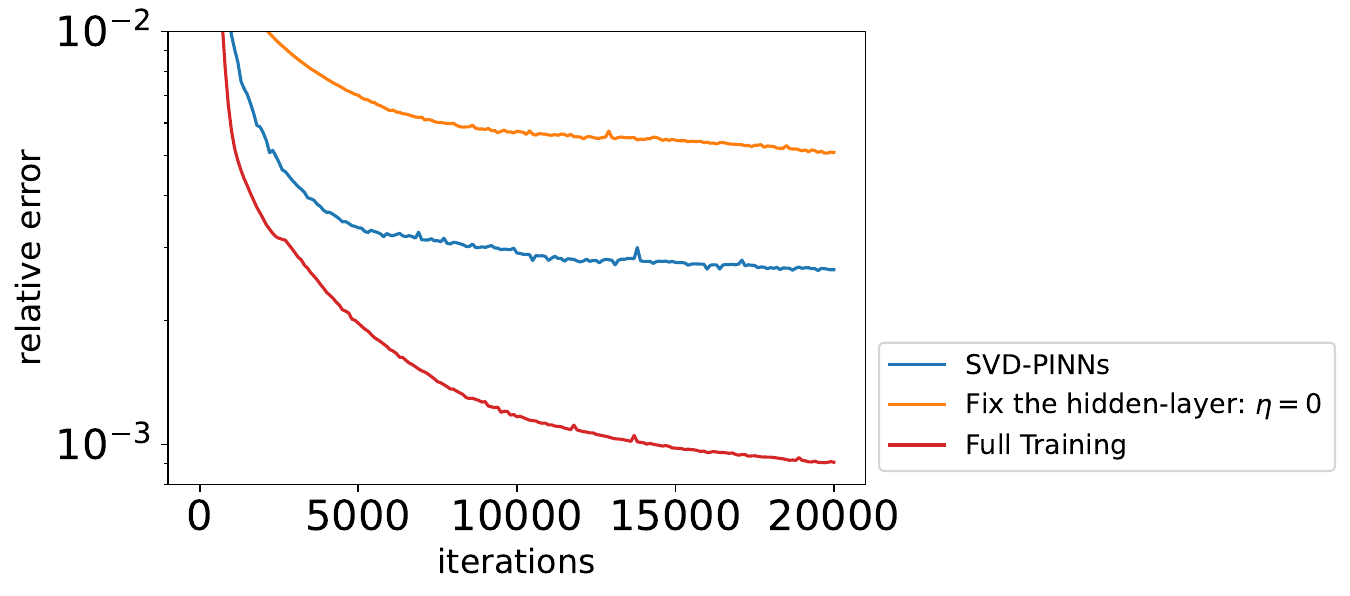}
    \caption{Comparisons of SVD-PINNs with the full training and the transfer learning with a frozen hidden layer in solving the $10$-dimensional Allen-Cahn equation ($\epsilon = 50$).}
    \label{allen_cahn_10d_error_50.00_comp}
\end{figure*}

\begin{figure*}[t!]
    \centering
    \includegraphics[width=0.63 \textwidth]{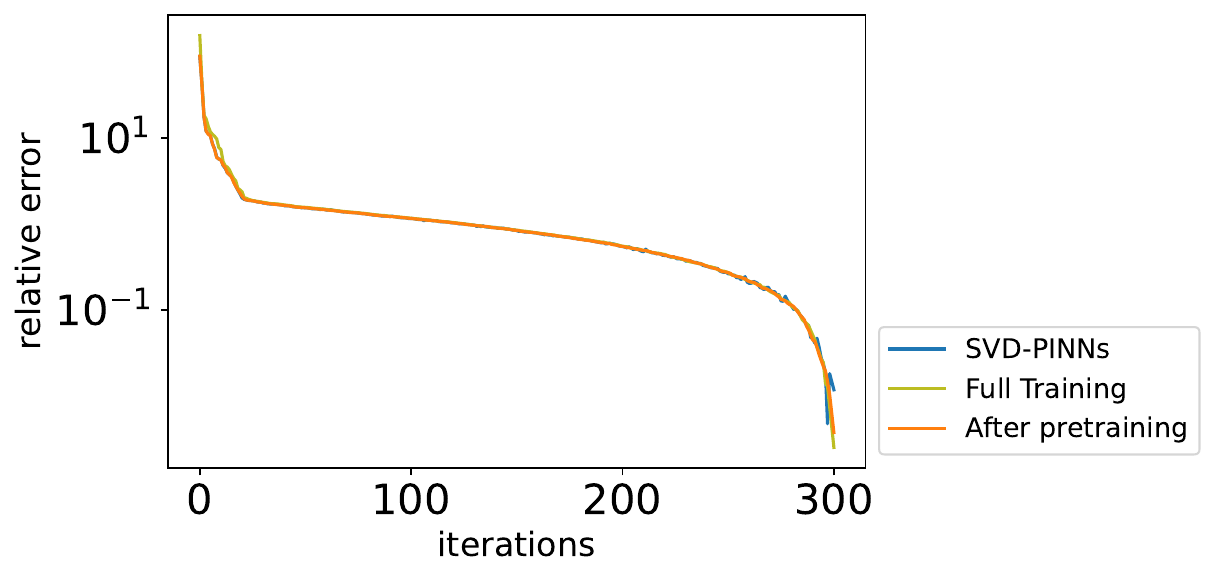}
    \caption{Singular values of $\mathbf{W}_1$ for the pretrained model, trained model with $\epsilon=0.5$.}
    \label{allen_cahn_10d_singular_values}
\end{figure*}

\begin{figure*}[t!]
    \centering
    \includegraphics[width=0.63 \textwidth]{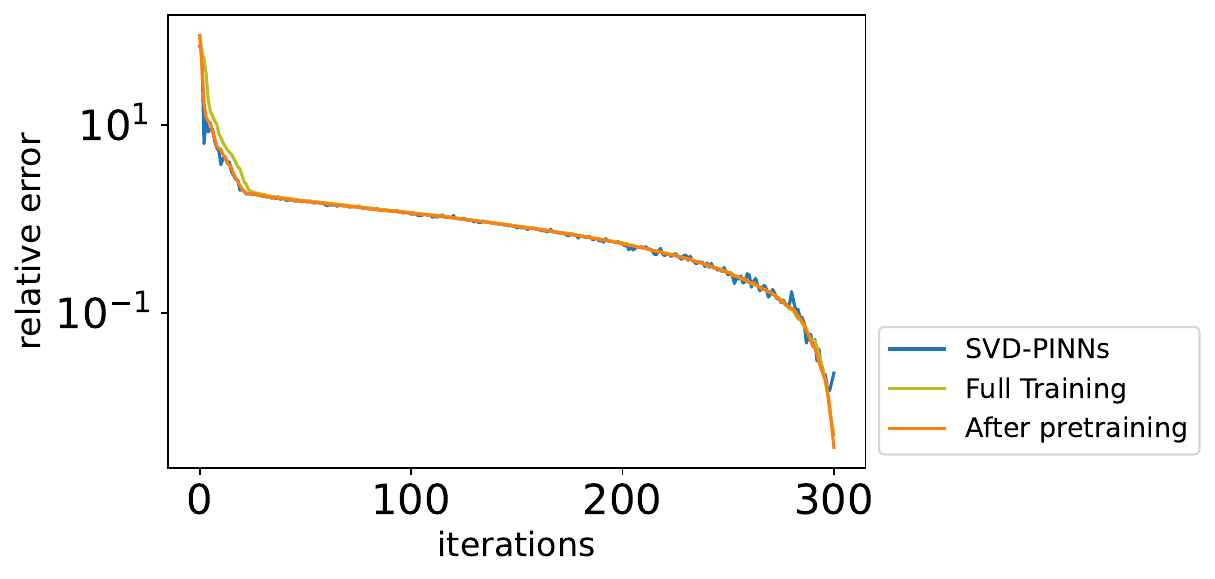}
    \caption{Singular values of $\mathbf{W}_1$ for the pretrained model, trained model with $\epsilon=50$.}
    \label{allen_cahn_10d_singular_values_50}
\end{figure*}

Results are displayed in Figures \ref{allen_cahn_10d_error_0.50} and \ref{allen_cahn_10d_error_2.00}. Here are our observations. Firstly, SVD-PINNs optimized by GD with learning rates $1$e-2 have the lowest relative error. However, in solving linear parabolic equations, RMSProp outperforms other optimizers. It further implies the difficulty of the optimization for singular values. We typically regard optimizers as hyperparameters in deep learning and choose them mainly by trials. In Figures \ref{allen_cahn_10d_error_0.50_comp} and \ref{allen_cahn_10d_error_2.00_comp}, it is within our expectation that the neural network with full training definitely achieves lower relative error if well optimized, because of its larger capacity compared to transfer learning models that fix some parameters. Figure \ref{allen_cahn_10d_error_50.00_comp} shows the effectiveness of training singular values in SVD-PINNs, especially for large $\epsilon$. In Figures \ref{allen_cahn_10d_error_0.50_comp}, \ref{allen_cahn_10d_error_2.00_comp} and \ref{allen_cahn_10d_error_50.00_comp}, the full training method achieves comparable results but the gap between SVD-PINNs and the simple transfer learning method is enlarged, in the case of large $\epsilon$. 
We also plot singular values of $\mathbf{W}_1$ for the pretrained model ($\epsilon=0$), fully trained models and SVD-PINNs with $\epsilon=0.5$ and $\epsilon=50$ in Figures \ref{allen_cahn_10d_singular_values} and \ref{allen_cahn_10d_singular_values_50}. We observed slight changes in singular values during training, even for large $\epsilon$. It further validates our conjecture that $\mathbf{W}_1$ are closely related to the differential operators.

\section{Conclusion and Discussion}
\label{sec_conclusion_discussion}
In this paper, we proposed a novel singular-values-based transfer learning of PINNs 
(SVD-PINNs), where singular vectors of the parameters matrix are frozen. Numerical investigations show that the transfer learning method stabilizes the training procedure compared with the full training. Moreover, singular-values optimization determines the performances of the SVD-PINNs. SVD-PINNs with suitable optimizers outperform the transfer learning method of PINNs in \cite{desai2021one} where $\mathbf{W}_1$ is frozen (cases with $\eta = 0$ in the numerical part).

Some future explorations are essential for the transfer learning of PINNs. Firstly, the theoretical analysis for the SVD-PINNs, e.g., the convergence and the generalization. Secondly, the optimization for singular values is challenging since it is a constrained optimization problem. Gradient-based (global) methods and projections are adopted for optimizing SVD-PINNs. However, they usually are not valid or not convergent in solving constrained optimization problems without the convexity condition. Numerical results show that successful optimization of singular values for SVD-PINNs contributes to the prediction of solutions, while SVD-PINNs with inaccurate singular values have large relative errors. Furthermore, the effectiveness of the SVD-PINNs and other transfer learning methods in solving PDEs with different but close differential operators or in other application backgrounds is an interesting topic.

\section*{Acknowledgement}
We would like to thank the anonymous reviewers for their helpful comments. This work is supported by Hong Kong Research Grant Council GRF 12300218, 12300519, 17201020, 17300021, C1013-21GF, C7004-21GF and Joint NSFC-RGC N-HKU76921.

\bibliographystyle{plain}
\bibliography{reference}

\begin{thebibliography}{10}

\bibitem{arjovsky2017wasserstein}
Martin Arjovsky, Soumith Chintala, and L{\'e}on Bottou.
\newblock Wasserstein generative adversarial networks.
\newblock In {\em International Conference on Machine Learning}, pages
  214--223. PMLR, 2017.

\bibitem{chen2021learning}
Xiaoli Chen, Jinqiao Duan, and George~Em Karniadakis.
\newblock Learning and meta-learning of stochastic
  advection--diffusion--reaction systems from sparse measurements.
\newblock {\em European Journal of Applied Mathematics}, 32(3):397--420, 2021.

\bibitem{desai2021one}
Shaan Desai, Marios Mattheakis, Hayden Joy, Pavlos Protopapas, and Stephen
  Roberts.
\newblock One-shot transfer learning of physics-informed neural networks.
\newblock {\em ICML AI4Science Workshop}, 2022.

\bibitem{gu2022deep}
Yiqi Gu and Michael~K Ng.
\newblock Deep neural networks for solving extremely large linear systems.
\newblock {\em arXiv preprint arXiv:2204.00313}, 2022.

\bibitem{hinton2012rmsprop}
Geoffrey Hinton.
\newblock Rmsprop: Divide the gradient by a running average of its recent
  magnitude.
\newblock {\em Neural Networks for Machine Learning}, 2012.

\bibitem{jagtap2020adaptive}
Ameya~D Jagtap, Kenji Kawaguchi, and George~Em Karniadakis.
\newblock Adaptive activation functions accelerate convergence in deep and
  physics-informed neural networks.
\newblock {\em Journal of Computational Physics}, 404:109136, 2020.

\bibitem{kingma2015adam}
Diederik~P Kingma and Jimmy Ba.
\newblock Adam: A method for stochastic optimization.
\newblock {\em International Conference for Learning Representations (ICLR)},
  2015.

\bibitem{krizhevsky2012imagenet}
Alex Krizhevsky, Ilya Sutskever, and Geoffrey~E Hinton.
\newblock Imagenet classification with deep convolutional neural networks.
\newblock {\em Advances in Neural Information Processing Systems}, 25, 2012.

\bibitem{lagaris1998artificial}
Isaac~E Lagaris, Aristidis Likas, and Dimitrios~I Fotiadis.
\newblock Artificial neural networks for solving ordinary and partial
  differential equations.
\newblock {\em IEEE Transactions on Neural Networks}, 9(5):987--1000, 1998.

\bibitem{lu2021learning}
Lu~Lu, Pengzhan Jin, Guofei Pang, Zhongqiang Zhang, and George~Em Karniadakis.
\newblock Learning nonlinear operators via deeponet based on the universal
  approximation theorem of operators.
\newblock {\em Nature Machine Intelligence}, 3(3):218--229, 2021.

\bibitem{miotto2018deep}
Riccardo Miotto, Fei Wang, Shuang Wang, Xiaoqian Jiang, and Joel~T Dudley.
\newblock Deep learning for healthcare: review, opportunities and challenges.
\newblock {\em Briefings in bioinformatics}, 19(6):1236--1246, 2018.

\bibitem{pan2009survey}
Sinno~Jialin Pan and Qiang Yang.
\newblock A survey on transfer learning.
\newblock {\em IEEE Transactions on Knowledge and Data Engineering},
  22(10):1345--1359, 2009.

\bibitem{pang2019fpinns}
Guofei Pang, Lu~Lu, and George~Em Karniadakis.
\newblock fpinns: Fractional physics-informed neural networks.
\newblock {\em SIAM Journal on Scientific Computing}, 41(4):A2603--A2626, 2019.

\bibitem{psichogios1992hybrid}
Dimitris~C Psichogios and Lyle~H Ungar.
\newblock A hybrid neural network-first principles approach to process
  modeling.
\newblock {\em AIChE Journal}, 38(10):1499--1511, 1992.

\bibitem{raissi2019physics}
Maziar Raissi, Paris Perdikaris, and George~E Karniadakis.
\newblock Physics-informed neural networks: A deep learning framework for
  solving forward and inverse problems involving nonlinear partial differential
  equations.
\newblock {\em Journal of Computational Physics}, 378:686--707, 2019.

\bibitem{shin2020convergence}
Yeonjong Shin, Jerome Darbon, and George~Em Karniadakis.
\newblock On the convergence of physics informed neural networks for linear
  second-order elliptic and parabolic type pdes.
\newblock {\em on Commun. Comput. Phys.}, (28):2042--2074, 2020.

\bibitem{sirignano2018dgm}
Justin Sirignano and Konstantinos Spiliopoulos.
\newblock Dgm: A deep learning algorithm for solving partial differential
  equations.
\newblock {\em Journal of computational physics}, 375:1339--1364, 2018.

\bibitem{vinyals2015grammar}
Oriol Vinyals, {\L}ukasz Kaiser, Terry Koo, Slav Petrov, Ilya Sutskever, and
  Geoffrey Hinton.
\newblock Grammar as a foreign language.
\newblock {\em Advances in Neural Information Processing Systems}, 28, 2015.

\bibitem{wang2022and}
Sifan Wang, Xinling Yu, and Paris Perdikaris.
\newblock When and why pinns fail to train: A neural tangent kernel
  perspective.
\newblock {\em Journal of Computational Physics}, 449:110768, 2022.

\bibitem{weinan2021algorithms}
E~Weinan, Jiequn Han, and Arnulf Jentzen.
\newblock Algorithms for solving high dimensional pdes: from nonlinear monte
  carlo to machine learning.
\newblock {\em Nonlinearity}, 35(1):278, 2021.

\bibitem{wong2021can}
Jian~Cheng Wong, Abhishek Gupta, and Yew-Soon Ong.
\newblock Can transfer neuroevolution tractably solve your differential
  equations?
\newblock {\em IEEE Computational Intelligence Magazine}, 16(2):14--30, 2021.

\bibitem{xu2020finite}
Jinchao Xu.
\newblock The finite neuron method and convergence analysis.
\newblock {\em arXiv preprint arXiv:2010.01458}, 2020.

\bibitem{yang2019adversarial}
Yibo Yang and Paris Perdikaris.
\newblock Adversarial uncertainty quantification in physics-informed neural
  networks.
\newblock {\em Journal of Computational Physics}, 394:136--152, 2019.

\bibitem{yosinski2014transferable}
Jason Yosinski, Jeff Clune, Yoshua Bengio, and Hod Lipson.
\newblock How transferable are features in deep neural networks?
\newblock {\em Advances in Neural Information Processing Systems}, 27, 2014.

\bibitem{zhang2020learning}
Dongkun Zhang, Ling Guo, and George~Em Karniadakis.
\newblock Learning in modal space: Solving time-dependent stochastic pdes using
  physics-informed neural networks.
\newblock {\em SIAM Journal on Scientific Computing}, 42(2):A639--A665, 2020.

\bibitem{zhang2019quantifying}
Dongkun Zhang, Lu~Lu, Ling Guo, and George~Em Karniadakis.
\newblock Quantifying total uncertainty in physics-informed neural networks for
  solving forward and inverse stochastic problems.
\newblock {\em Journal of Computational Physics}, 397:108850, 2019.

\end{thebibliography}
\end{document}